\documentclass[conference]{IEEEtran}
\IEEEoverridecommandlockouts
\usepackage{cite}
\usepackage{amsmath,amssymb,amsfonts}
\usepackage{algorithmic}
\usepackage{graphicx}
\usepackage{textcomp}
\usepackage{xcolor}
\def\BibTeX{{\rm B\kern-.05em{\sc i\kern-.025em b}\kern-.08em
    T\kern-.1667em\lower.7ex\hbox{E}\kern-.125emX}}

\usepackage{enumitem}

\usepackage{hyperref}
\usepackage{booktabs}
\begin{document}

\title{
ByteScience: Bridging Unstructured Scientific Literature and Structured Data with Auto Fine-tuned Large Language Model in Token Granularity
}

\author{
    \IEEEauthorblockN{Tong Xie\IEEEauthorrefmark{1}\IEEEauthorrefmark{5}, Hanzhi Zhang\IEEEauthorrefmark{2}, Shaozhou Wang\IEEEauthorrefmark{1}, Yuwei Wan\IEEEauthorrefmark{1}, Imran Razzak\IEEEauthorrefmark{3}, \\ Chunyu Kit\IEEEauthorrefmark{4}, Wenjie Zhang\IEEEauthorrefmark{3}, and Bram Hoex\IEEEauthorrefmark{5}}
    
    \IEEEauthorblockA{\IEEEauthorrefmark{1}GreenDynamics, Kensington, Australia \\
    \{tong, shaozhou, yuwei\}@greendynamics.com.au}
    
    \IEEEauthorblockA{\IEEEauthorrefmark{2}Dept. of Computer Science and Engineering, University of North Texas, Denton, United States, hanzhizhang@my.unt.edu}
    
    \IEEEauthorblockA{\IEEEauthorrefmark{3}School of Computer Science and Engineering, University of New South Wales, Kensington, Australia \\
    \{imran.razzak, wenjie.zhang\}@unsw.edu.au}
    
    \IEEEauthorblockA{\IEEEauthorrefmark{4}Dept. of Linguistics and Translation, City University of Hong Kong, Hong Kong, China, ctckit@cityu.edu.hk}
    
    \IEEEauthorblockA{\IEEEauthorrefmark{5}School of Photovoltaic and Renewable Energy Engineering, University of New South Wales, Kensington, Australia \\
    \{tong, b.hoex\}@unsw.edu.au}
}

\IEEEaftertitletext{\vspace{-2em}}

\maketitle

\begin{abstract}
Natural Language Processing (NLP) is widely used to supply summarization ability from long context to structured information.  
However, extracting structured knowledge from scientific text by NLP models remains a challenge because of its domain-specific nature to complex data preprocessing and the granularity of multi-layered device-level information.
To address this, we introduce \href{http://byte-science.com}{ByteScience}, a non-profit cloud-based auto fine-tuned Large Language Model (LLM) platform, which is designed to extract structured scientific data and synthesize new scientific knowledge from vast scientific corpora. 
The platform capitalizes on DARWIN, an open-source, fine-tuned LLM dedicated to natural science. The platform was built on Amazon Web Services (AWS) and provides an automated, user-friendly workflow for custom model development and data extraction. 
The platform achieves remarkable accuracy with only a small amount of well-annotated articles. This innovative tool streamlines the transition from the science literature to structured knowledge and data and benefits the advancements in natural informatics. \href{https://youtu.be/S4RHt67f4pA}{Demo Video}
\end{abstract}


\vspace{-2pt}
\section{Introduction}

AI has the potential to revolutionize scientific discovery (AI4Science \cite{stevens_ai_2020}), but challenges remain. Scientific knowledge is scattered across documents, making it hard to fully leverage past research. LLMs offer a promising solution but require structured texts, including converting PDFs and generating fine-tuning examples for NLP tasks. While machine learning models are used in fields like drug discovery \cite{vamathevan_applications_2019}, protein design \cite{noauthor_machine_nodate}, and crystal structure generation \cite{xie_opinion_2024}, limited structured data hinders their effectiveness. Databases like Materials Project \cite{jain_commentary_2013} and NOMAD \cite{draxl_nomad_2019} cover only a fraction of data, leaving much unstructured information untapped. This gap presents an opportunity for AI to accelerate discovery. Although converting documents to markup is well-studied, extracting complex relationships remains challenging but essential for building knowledge graphs and fine-tuning datasets.

\begin{itemize}[leftmargin=0.3cm]
    \setlength{\itemsep}{0pt} 
    \setlength{\parskip}{0pt} 
    \item \textbf{Contextual Dependency}: Relationships in scientific texts often depend heavily on context that may span multiple sentences or sections. For instance, a material's properties might be discussed concerning its synthesis method, as described in paragraphs several pages before.
    \item \textbf{Implicit Connections}: Many relationships in scientific writing are implied rather than explicitly stated, requiring deep domain knowledge to infer correctly.
    \item \textbf{Hierarchical Structures}: Scientific documents frequently contain nested relationships, such as experiment subsets or multi-step processes, which are challenging to represent in flat data structures.
    \item \textbf{Cross-Reference Complexity}: Relationships often span different document parts, such as tables, figures, and citations, requiring holistic understanding.
    \item \textbf{Domain-Specific Semantics}: Each scientific field has unique terminology and conventions, complicating universal extraction methods. For example, pseudocode and flowcharts may be unfamiliar in other domains.
\end{itemize}

Traditional methods like MatKG \cite{venugopal_matkg_2022}, which define relationships by entity co-occurrence, often miss the nuances of scientific knowledge. While useful, they risk oversimplifying complex relationships. Advanced techniques are needed to better capture this complexity for improved knowledge extraction in AI-driven scientific discovery. Therefore, we introduce \href{http://byte-science.com}{ByteScience}, a cloud-based platform featuring an auto-fine-tuned LLM to extract structured scientific data and synthesize new scientific knowledge from extensive scientific corpora. We conclude as follows:
\begin{enumerate}[leftmargin=0.4cm]
    \setlength{\itemsep}{0pt} 
    \setlength{\parskip}{0pt} 
    \item Tailored with DARWIN \cite{xie_darwin_2023}, an open-source state-of-the-art nature-science LLM, to provide research focus utilization;
    \item Zero-code user-friendly semi-automated annotation and processing for uploaded science documents;
    \item A personalized and domain-specific auto fine-tuning LLM that requires only a single fully annotated piece of literature;
    \item Time efficiency high-quality science data extraction from millions of papers for less than a second per article.
\end{enumerate}

\begin{figure*}[t]
  \includegraphics[width=1\linewidth]{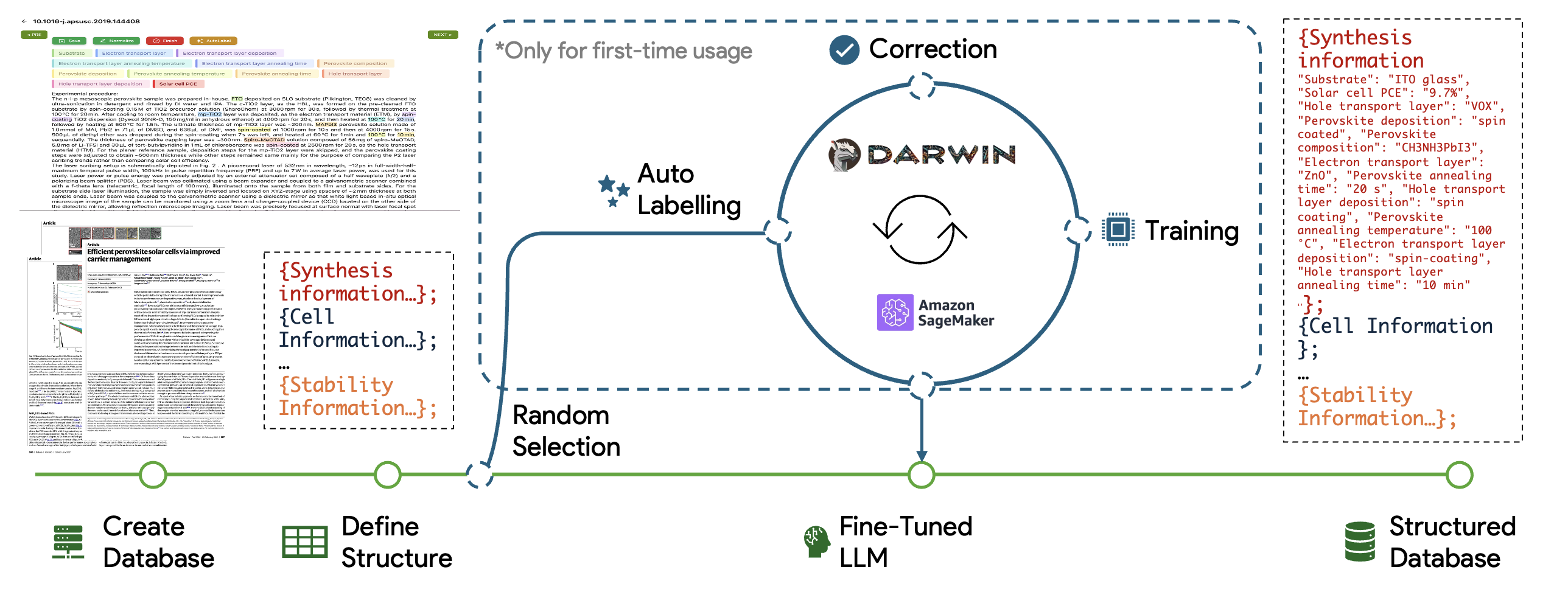}
  \caption {\textbf{ByteScience Pipeline.} The initial setup for a specific field involves constructing a domain-specific corpus of structured scientific data (Green Pipeline) and fine-tuning an LLM on this dataset to optimize performance for the target scientific domain (Blue Pipeline). Once this setup is complete, users can efficiently generate structured datasets from new scientific documents in the same field by utilizing the fine-tuned LLM stored in AWS.}
  \label{fig1}  
\end{figure*}

\vspace{-2pt}
\section{Platform Design}
ByteScience is a robust, scalable cloud-based solution leveraging AWS Sagemaker. This architecture ensures high availability, scalability, and performance for processing large scientific documents. Figure~\ref{fig1} illustrates the extraction pipeline for custom model development and data extraction. The pipeline consists of two primary phases:

\begin{itemize}[leftmargin=0.3cm]
    \setlength{\itemsep}{0pt} 
    \setlength{\parskip}{0pt} 
    \item \textbf{Initial Setup (First-time use for a specific field):}
    
    \textit{Dataset Construction (Green Pipeline):} This phase builds a domain-specific corpus of structured scientific data.
    
    \textit{LLM Fine-tuning (Blue Pipeline):} The system fine-tunes a large language model on the constructed dataset to optimize performance for the target scientific domain.
    
    \item \textbf{Operational Phase:}
    Once the initial setup is complete, users can directly utilize the fine-tuned LLM stored in AWS to efficiently generate structured datasets from new scientific documents in the same field.
\end{itemize}

This two-phase approach allows ByteScience to quickly adapt to various scientific domains while maintaining high extraction accuracy. The cloud-based architecture enables seamless scaling and ensures users always have access to the latest fine-tuned models, streamlining the conversion of unstructured scientific literature into structured data. Key steps include:
\begin{enumerate}[leftmargin=0.4cm]
\setlength{\itemsep}{0pt} 
\setlength{\parskip}{0pt} 
\item \textbf{Create Database:}
Users upload scientific documents in JSON, PDF, HTML, or XML formats. Non-JSON text is extracted and saved as JSON, with HTML/XML markup stripped, and PDF conversion done using PDFMiner \cite{shinyama_pdfminer_2015}.
\item \textbf{Define Structure:}
Users define annotation structures, including entity labels and relationships, using pre-built or custom templates.
\item \textbf{Random Selection:}
A small text subset is randomly selected for initial annotation on first use.
\item \textbf{Auto Labelling:}
The LLM applies automatic pre-labeling to the selected texts.
\item \textbf{Correction:}
Users review and correct the auto-labeled annotations, ensuring accuracy and consistency.
\item \textbf{Training:}
Corrected annotations are used to train or fine-tune an LLM, with training done via Amazon SageMaker.
\item \textbf{Fine-Tuned LLM:}
The training process results in a fine-tuned LLM customized for the specific annotation task.
\item \textbf{Structured Data Generation:} The fine-tuned LLM processes new documents into structured data stored in MongoDB as JSON, allowing flexible use and efficient querying.
\end{enumerate}

\begin{figure*}[t]
\centering
  \includegraphics[width=0.7\linewidth]{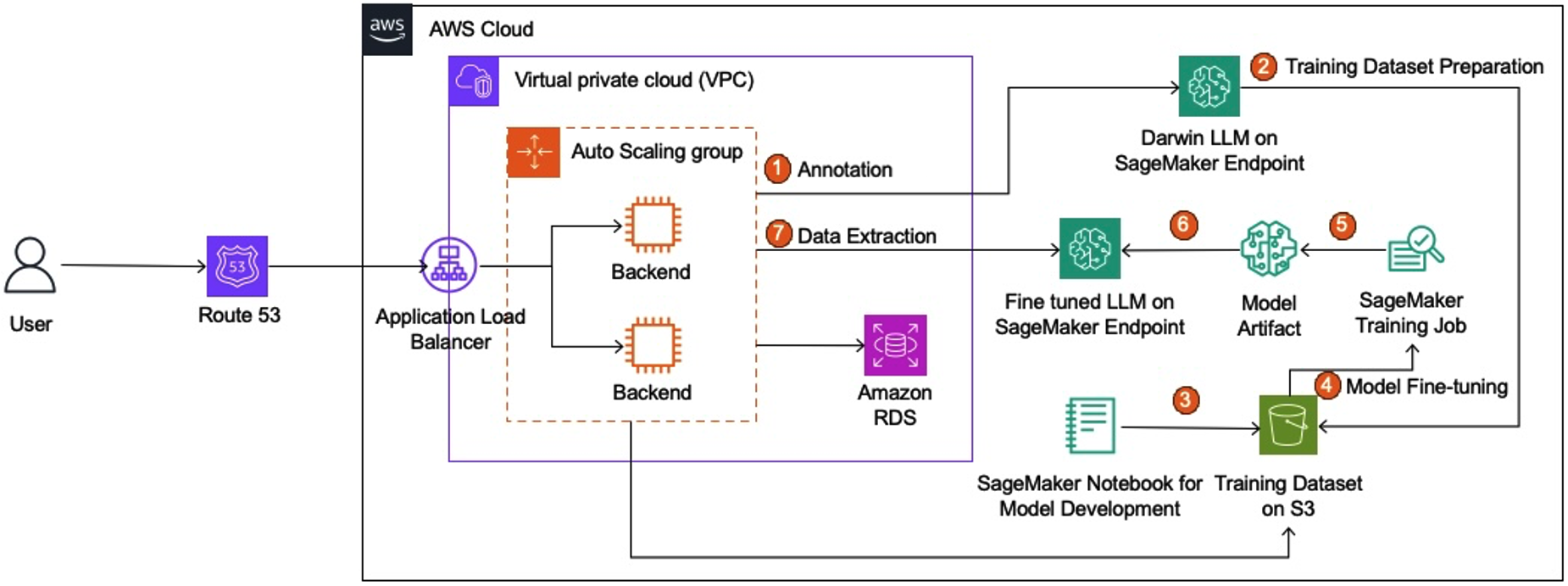}
  \caption {The architecture of ByteScience creates a structured database on AWS cloud with LLM.}
  \label{fig2}  
\end{figure*}

After uploading the training dataset, it is transformed into the LLM’s instruction format for fine-tuning. Users should first test a small text subset and assess accuracy and recall. If accuracy is low, add more annotated data and retrain. For low recall, generate more corpus data and use sequential learning to train a new model.

\vspace{-2pt}
\section{Architecture of AWS Cloud-Based Services}

ByteScience utilizes the robust, scalable infrastructure of Amazon Web Services (AWS) to efficiently handle user requests and data processing. Figure~\ref{fig2} shows the detailed architecture of our platform.

\subsection{General Service(Green Pipeline) Infrastructure}

The user interaction layer is built on a series of AWS services that ensure high availability, security, and performance:

\begin{itemize}[leftmargin=0.3cm]
    \setlength{\itemsep}{0pt} 
    \setlength{\parskip}{0pt} 
    \item \textbf{DNS Management:} AWS Route 53 routes incoming user requests to the appropriate services within the architecture.
    
    \item \textbf{Load Balancing:} An Application Load Balancer (ALB) distributes traffic evenly across multiple backend servers, ensuring fault tolerance and optimal performance.
    
    \item \textbf{Compute Resources:} Backend servers, organized in an Auto Scaling group within a Virtual Private Cloud (VPC), provide a secure, scalable environment to traffic demands.
    
    \item \textbf{Database Services:} Amazon Relational Database Service (RDS) supports complex queries and transactions essential for scientific data management.
\end{itemize}


\subsection{LLM Service (Blue Pipeline) Architecture}

The LLM fine-tuning capability is a core function, implemented through a sophisticated pipeline of AWS services:
\begin{itemize}[leftmargin=0.3cm]
    \setlength{\itemsep}{0pt} 
    \setlength{\parskip}{0pt} 
    \item \textbf{Model Development:} SageMaker Notebook is used for developing foundation models. We reproduced the DARWIN model by fine-tuning LLaMA 7B, leveraging SageMaker’s computational tools.
    
    \item \textbf{a DARWIN LLM processes Training Dataset Preparation:} User annotations hosted on a SageMaker Endpoint, preparing data for model training.
    
    \item \textbf{Data Storage:} Training datasets are securely stored on Amazon S3, ensuring durability and accessibility.
    
    \item \textbf{Model Fine-tuning:} A SageMaker Training Job handles fine-tuning the model at scale, utilizing AWS's distributed computing capabilities.
    
    \item \textbf{Model Deployment:} The fine-tuned model is deployed to a SageMaker Endpoint, providing a managed environment for querying the LLM for data extraction tasks.
\end{itemize}

\subsection{Workflow Integration}

ByteScience workflow seamlessly integrates these components:

\begin{enumerate}[leftmargin=0.4cm]
    \setlength{\itemsep}{0pt} 
    \setlength{\parskip}{0pt} 
    \item Users interact with the system through Route 53 and the ALB for initial annotation tasks.
    \item The DARWIN LLM processes annotated data on a SageMaker Endpoint.
    \item The SageMaker Notebook is used for model development and improvement.
    \item Training datasets on S3 are used to fine-tune the model via SageMaker Training Jobs.
    \item The resulting model is deployed to a SageMaker Endpoint.
    \item Users can then perform data extraction tasks, processing requests by the fine-tuned LLM.
\end{enumerate}

This architecture enables ByteScience to offer customized, high-performance language models tailored to specific scientific domains, facilitating accurate and efficient structured data extraction from unstructured scientific literature.

\vspace{-2pt}
\section{Structured Data Extraction Performance}
LLMs significantly improve human-in-the-loop annotation. Using 300 training samples reduced annotation time by 57\% compared to a single sample \cite{dunn_structured_2022}. In the GPT-3/Doping-English model, 10-20 samples were enough to learn the correct structure, with precision, recall, and F1 scores reaching 0.8-0.9 with around 300 samples.

In our experiment, we compared non-LLM and LLM methods for structured data extraction on 90 samples covering batteries, catalysis, and photovoltaics, alongside ByteScience's results. As shown in Table \ref{table:Result of NER, RE and ER through Fine-tuned LLMs}, we evaluated Named Entity Recognition (NER), Relation Extraction (RE), and Entity Resolution (ER). While models like MatBERT performed well, they often produced irrelevant entities, lowering precision. In contrast, LLMs handled unstructured information more reliably, and our system outperformed traditional methods across all tasks with fewer samples.

\begin{table}[h!]
\caption{Result of structured data extraction.}
  \centering
  \scriptsize
  \begin{tabular}{ccccc}
    \toprule
    \textbf{Task} & \textbf{Model} & \textbf{Precision} & \textbf{Recall} & \textbf{F1 score} \\
    \midrule
    & MatBERT & \textcolor{gray}{0.1196} & \textcolor{gray}{0.6869} & \textcolor{gray}{0.2036} \\
    & Llama 7b & \textcolor{gray}{0.6101} & \textcolor{gray}{0.6216} & \textcolor{gray}{0.6158} \\
    NER & Llama2 7b & \textcolor{gray}{0.7419} & \textcolor{gray}{0.7667} & \textcolor{gray}{0.7541} \\
    & Darwin & \textcolor{gray}{0.8013} & \textcolor{gray}{0.7935} & \textcolor{gray}{0.7974} \\
    & Bytescience & 0.9520 & 0.9083 & 0.9296 \\
    \midrule
    & MatBERT & \textcolor{gray}{0.0250} & \textcolor{gray}{0.5696} & \textcolor{gray}{0.0479} \\
    & Llama 7b & \textcolor{gray}{0.5305} & \textcolor{gray}{0.5405} & \textcolor{gray}{0.5355} \\
    RE & Llama2 7b & \textcolor{gray}{0.6452} & \textcolor{gray}{0.6667} & \textcolor{gray}{0.6557} \\
    & Darwin & \textcolor{gray}{0.7036} & \textcolor{gray}{0.6968} & \textcolor{gray}{0.7002} \\
    & Bytescience & 0.9039 & 0.8625 & 0.8827 \\
    \midrule
    & MatBERT & \textcolor{gray}{0.0928} & \textcolor{gray}{0.5303} & \textcolor{gray}{0.1579} \\
    & Llama 7b & \textcolor{gray}{0.3687} & \textcolor{gray}{0.3757} & \textcolor{gray}{0.3722} \\
    ER & Llama2 7b & \textcolor{gray}{0.4484} & \textcolor{gray}{0.4633} & \textcolor{gray}{0.4557} \\
    & Darwin & \textcolor{gray}{0.4593} & \textcolor{gray}{0.4548} & \textcolor{gray}{0.4571} \\
    & Bytescience & 0.9127 & 0.8708 & 0.8913 \\
    \bottomrule
  \end{tabular}
  \label{table:Result of NER, RE and ER through Fine-tuned LLMs}
\end{table}

\vspace{-2pt}
\section{ByteScience in Action: A User Case Study}

To showcase ByteScience's application, we present Thomas, a materials scientist automating alloy synthesis by analyzing literature to establish "Composition-Processing-Structure-Performance" (CPSP) relationships. He designs alloy compositions, develops processing methods, and predicts microstructures using data on casting, solution treatment, and aging.

\subsection{Initial Setup: Schema, Semi-Annotation, Model Fine-Tune}

Thomas configures ByteScience to meet his research needs by designing a custom annotation schema for alloy synthesis, annotating key details like compositions, casting parameters, solution treatment, and aging variables. ByteScience then initiates semi-automatic annotation, where the DARWIN LLM auto-labels papers from his corpus based on this schema. Thomas reviews and corrects the annotations to refine the model's understanding. Afterward, ByteScience fine-tunes the LLM using AWS SageMaker, optimizing it for alloy synthesis data extraction. The fine-tuned model is deployed to a SageMaker Endpoint for efficient, large-scale processing of complex scientific papers.

\subsection{Data Generation: Document Upload, Endpoint Utilization, and Dataset Creation}

With the fine-tuned model, Thomas uploads his entire corpus of scientific papers to ByteScience, which processes various formats for comprehensive coverage. He initiates large-scale data extraction via the SageMaker Endpoint, where the model extracts detailed information on alloy compositions, casting processes, solution treatments, and aging procedures. This automation accelerates his research, completing in days what would have taken months manually. The extracted data is structured and stored in MongoDB, allowing Thomas to easily query, analyze, and identify trends in alloy synthesis, uncovering insights that manual review might have missed.

\subsection{Further Dataset Updates and Refinement}

As Thomas advances in his research, he updates his dataset with ByteScience, uploading new papers and processing them through the fine-tuned model to continually enrich his dataset. When discrepancies or improvements are needed, he initiates a re-training cycle, reviewing and correcting a subset of the newly processed papers to further fine-tune the model. This iterative process ensures the model stays accurate and adapts to evolving terminologies or methods in alloy synthesis. Through this dynamic interaction, Thomas maintains an up-to-date, accurate dataset, enhancing his research and keeping him at the forefront of alloy synthesis advancements.

\vspace{-2pt}
\section{Significance to Science}

Constructing databases from scholarly literature is crucial for modern research, but traditional methods are time-consuming and resource-intensive. ByteScience transforms this process by enabling users to create a customized data extraction tool in hours, achieving 80\%-90\% human accuracy. It can process a 10-page scientific document in one second, compared to the 20-30 minutes it takes a researcher. With an extraction cost of just \$0.023 per paper for 10,000 articles, ByteScience makes large-scale data extraction affordable and accessible. Its versatility across scientific fields democratizes access to advanced data extraction, providing computational power equivalent to hundreds of annotators. This accelerates discovery, enhances research decision-making, and fosters innovation across disciplines.

\vspace{-2pt}
\section{Conclusion}


ByteScience is leveraging a powerful approach to handle unstructured text by fine-tuning DARWIN, a pre-trained natural science LLM, using a minimal set of annotated articles. Hosted on the AWS cloud, this platform automates the process of extracting structured data from scientific texts, presenting a zero-code solution that could significantly enhance efficiency in natural science research.
The key advantage of ByteScience lies in its ability to train the DARWIN model with few annotations, making it exceptionally adaptive and efficient. This capability ensures that the extracted material data is high-quality and highly accurate. ByteScience exemplifies how cutting-edge technology can be harnessed to propel advancements in science, engineering, and research by integrating advanced NLP techniques with cloud computing. This initiative represents a substantial step forward in making vast scientific corpora more accessible and usable, highlighting the transformative potential of AI in scientific data processing.
To optimize resource efficiency, we are developing a slicing version that fine-tunes a low-resource inference model using only partial data from extensive content.

\bibliography{custom}

\appendix

\begin{figure}[h]
\centering
  \includegraphics[width=0.5\textwidth]{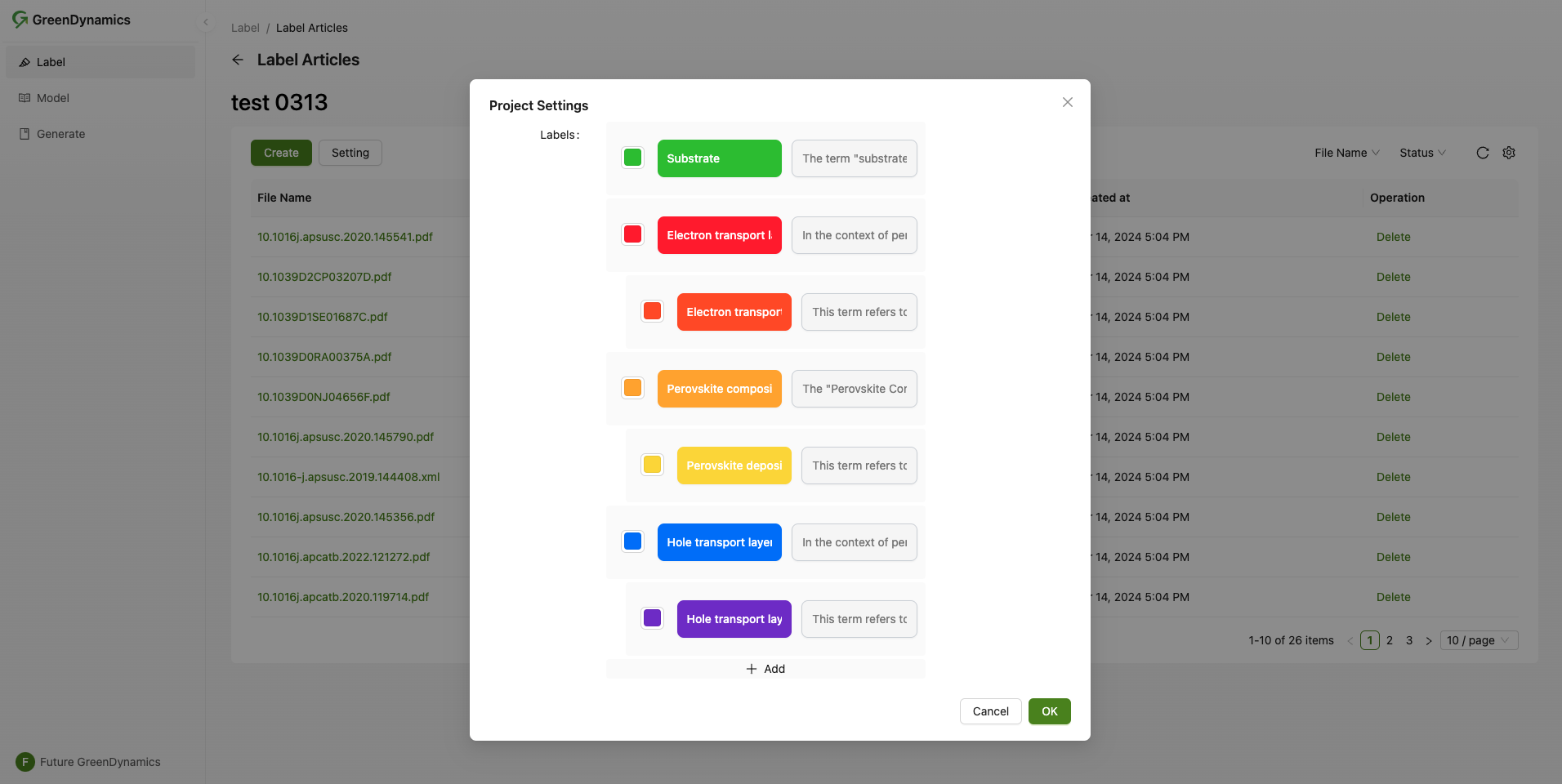}
  \caption{Screenshot of label setup. }
\label{src1}
\end{figure}

\autoref{src1} shows the label-defining function in our system. Users define the structure for annotations, including entity labels and their relationships (visualized by indent). For each label, there is a definition textbox for filling.

\begin{figure}[h]
\centering
  \includegraphics[width=0.5\textwidth]{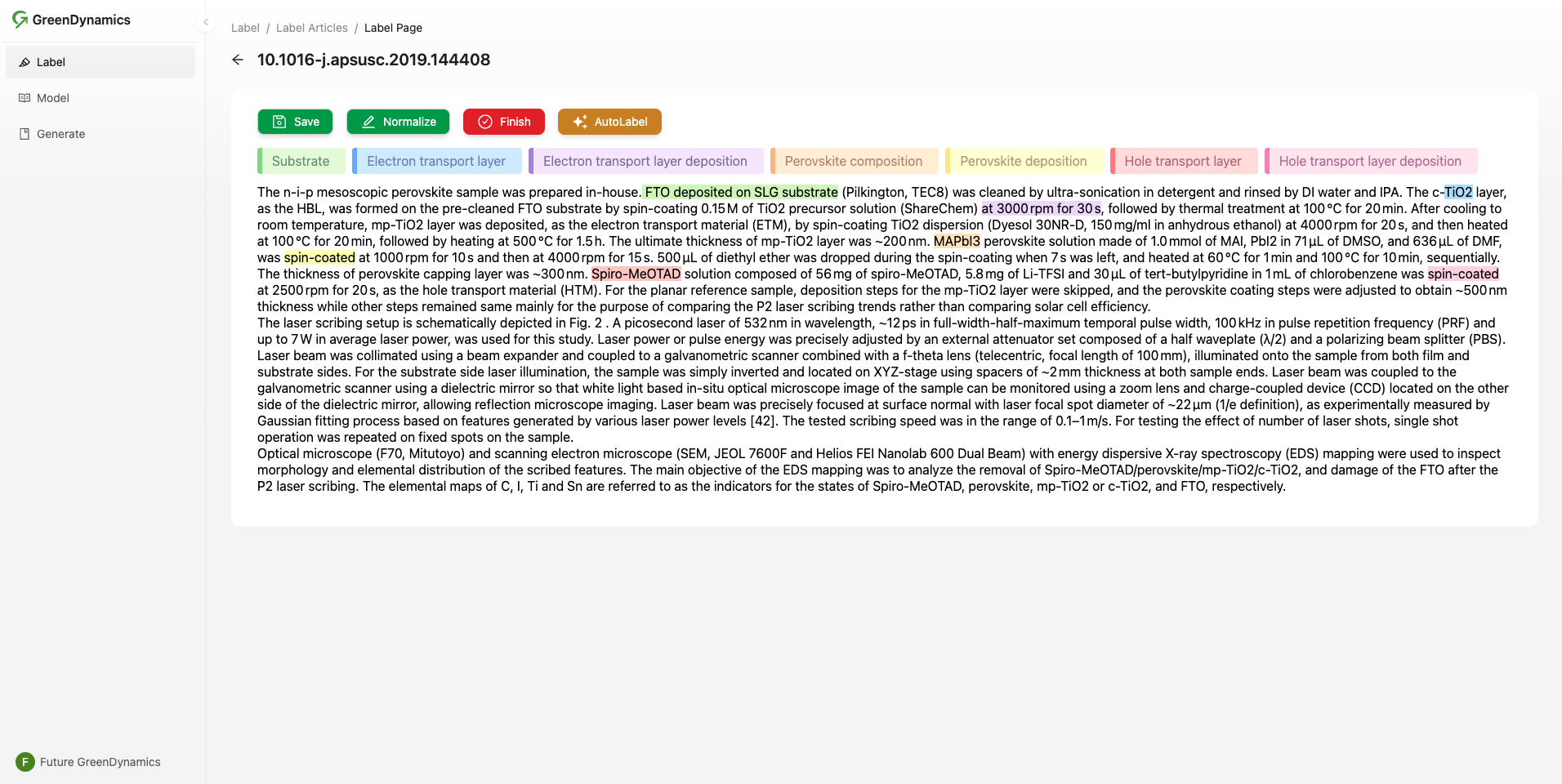}
  \caption{Screenshot of labeling page. }
\label{src2}
\end{figure}

\autoref{src2} shows the annotation function. Users can use automatic pre-labelling to the selected texts. Different colors will visualize the auto-labeled annotations and users can review and correct them, ensuring accuracy and consistency.

\begin{figure}[h]
\centering
  \includegraphics[width=0.5\textwidth]{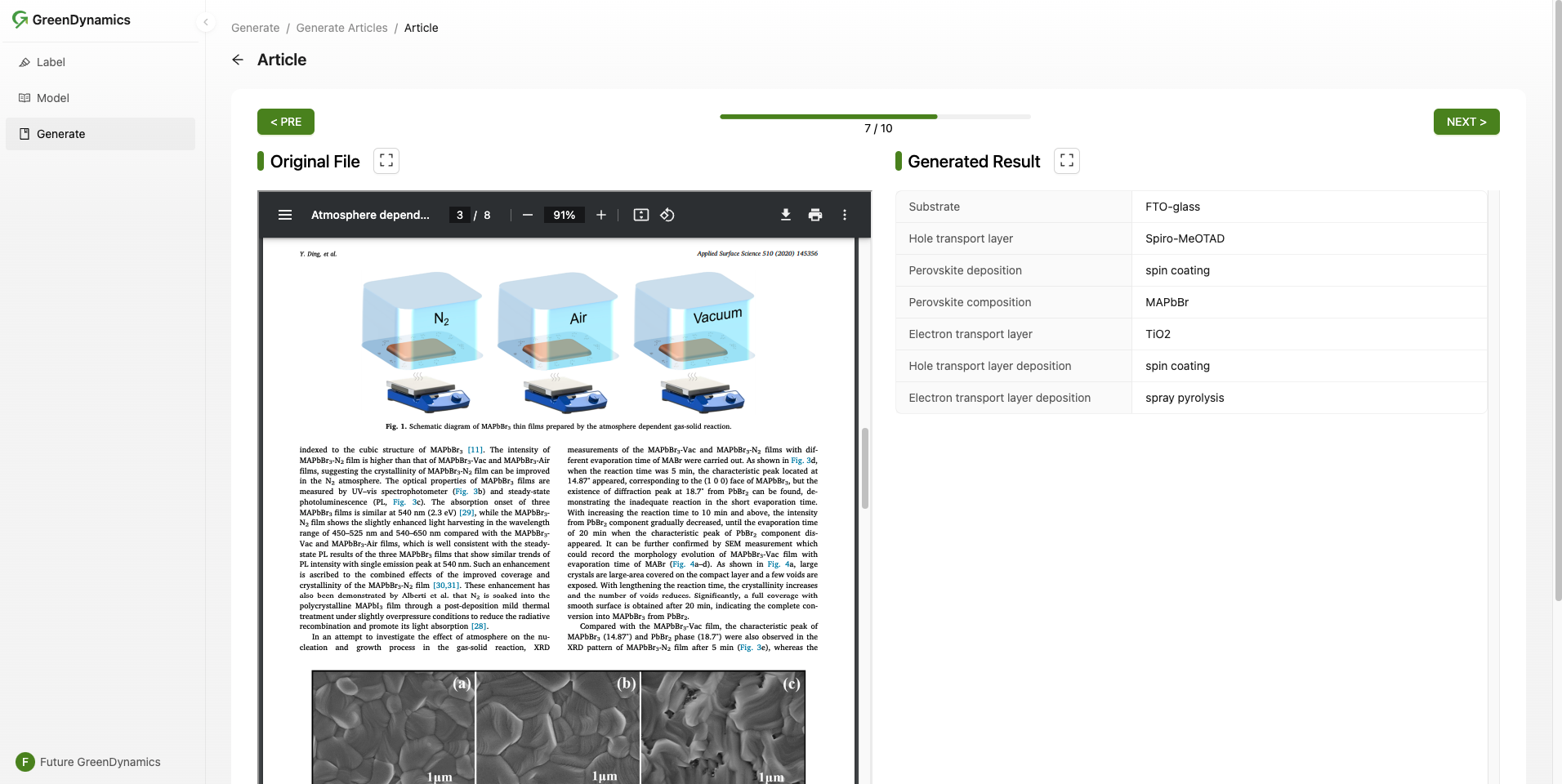}
  \caption{Screenshot of extraction results of a paper. }
\label{src3}
\end{figure}

\autoref{src3} shows the data extraction function on example paper. Users can create a customized data extraction tool that achieves 80\%-90\% of human extraction accuracy after just a few hours of annotation.

\end{document}